\documentclass[conference]{IEEEtran}
\IEEEoverridecommandlockouts
\usepackage{cite}
\usepackage{amsmath,amssymb,amsfonts}
\usepackage{algorithmic}
\usepackage{graphicx}
\usepackage{textcomp}
\usepackage{xcolor}

\usepackage{graphicx}
\usepackage{booktabs}
\usepackage{colortbl}
\usepackage{amsthm,amsmath,amssymb}
\usepackage{mathrsfs}
\usepackage{multirow}
\usepackage{makecell}
\usepackage{bbm}

\usepackage[pagebackref,breaklinks,colorlinks]{hyperref}

\def\BibTeX{{\rm B\kern-.05em{\sc i\kern-.025em b}\kern-.08em
    T\kern-.1667em\lower.7ex\hbox{E}\kern-.125emX}}
\begin{document}

\title{Exploring Part-Informed Visual-Language Learning for Person Re-Identification}

\author{Yin Lin$^{1,2}$, Yehansen Chen$^{2}$, Baocai Yin$^{2}$, Jinshui Hu$^{2}$, Bing Yin$^{2}$, Cong Liu$^{2}$, Zengfu Wang$^{1}$\footnotemark*\\
$^{1}$ University of Science and Technology of China, Hefei, China\\
$^{2}$ iFLYTEK Research, Hefei, China\\
{\tt\small lin5875@mail.ustc.edu.cn; zfwang@ustc.edu.cn; } \\
{\tt\small\{yinlin, yhschen, bcyin, jshu, bingyin, congliu2\}@iflytek.com}}

\maketitle
\renewcommand{\thefootnote}{$*$}
\footnotetext[1]{Corresponding author}

\begin{abstract}
  Recently, visual-language learning (VLL) has shown great potential in enhancing visual-based person re-identification (ReID). 
  Existing VLL-based ReID methods typically focus on image-text feature alignment at the whole-body level, while neglecting supervision on fine-grained part features, thus lacking constraints for local feature semantic consistency.
  To this end, we propose \textbf{P}art-Informed \textbf{V}isual-language \textbf{L}earning ($\pi$-VL) to enhance fine-grained visual features with part-informed language supervisions for ReID tasks. Specifically, $\pi$-VL introduces a human parsing-guided prompt tuning strategy and a hierarchical visual-language alignment paradigm to ensure within-part feature semantic consistency. The former combines both identity labels and human parsing maps to constitute pixel-level text prompts, and the latter fuses multi-scale visual features with a light-weight auxiliary head to perform fine-grained image-text alignment. As a plug-and-play and inference-free solution, our $\pi$-VL achieves performance comparable to or better than state-of-the-art methods on four commonly used ReID benchmarks. Notably, it reports \textbf{91.0\%} Rank-1 and \textbf{76.9\%} mAP on the challenging MSMT17 database, without bells and whistles.
\end{abstract}

\begin{IEEEkeywords}
Person re-identification, Visual-language learning, Fine-grained image-text alignment
\end{IEEEkeywords}

\section{Introduction}
\label{sec:intro}
Person re-identification (ReID) refers to the task of retrieving the query person-of-interest from large-scale gallery databases captured by non-overlapping camera views \cite{ye2021deep}. Owing to its practical importance for intelligent video surveillance, ReID has gained ever-growing attention from both academia and industry in recent years \cite{Luo_2019_CVPR_Workshops, Sun_2018_ECCV, wang2018learning}.

As appearance biometrics serve as the most fundamental and well-studied cues for identity recognition \cite{quan2019auto, zhou2019omni, he2021transreid, Luo_2019_CVPR_Workshops}, appearance-based ReID has achieved considerable success across a wide range of applications. However, human body semantics are not readily apparent in raw pixels, making it challenging to learn semantic information under the single supervision of one-hot or pair-wise identity labels \cite{li2023clip}.


\begin{figure}[t]
\begin{center}
   \includegraphics[width=0.8\linewidth]{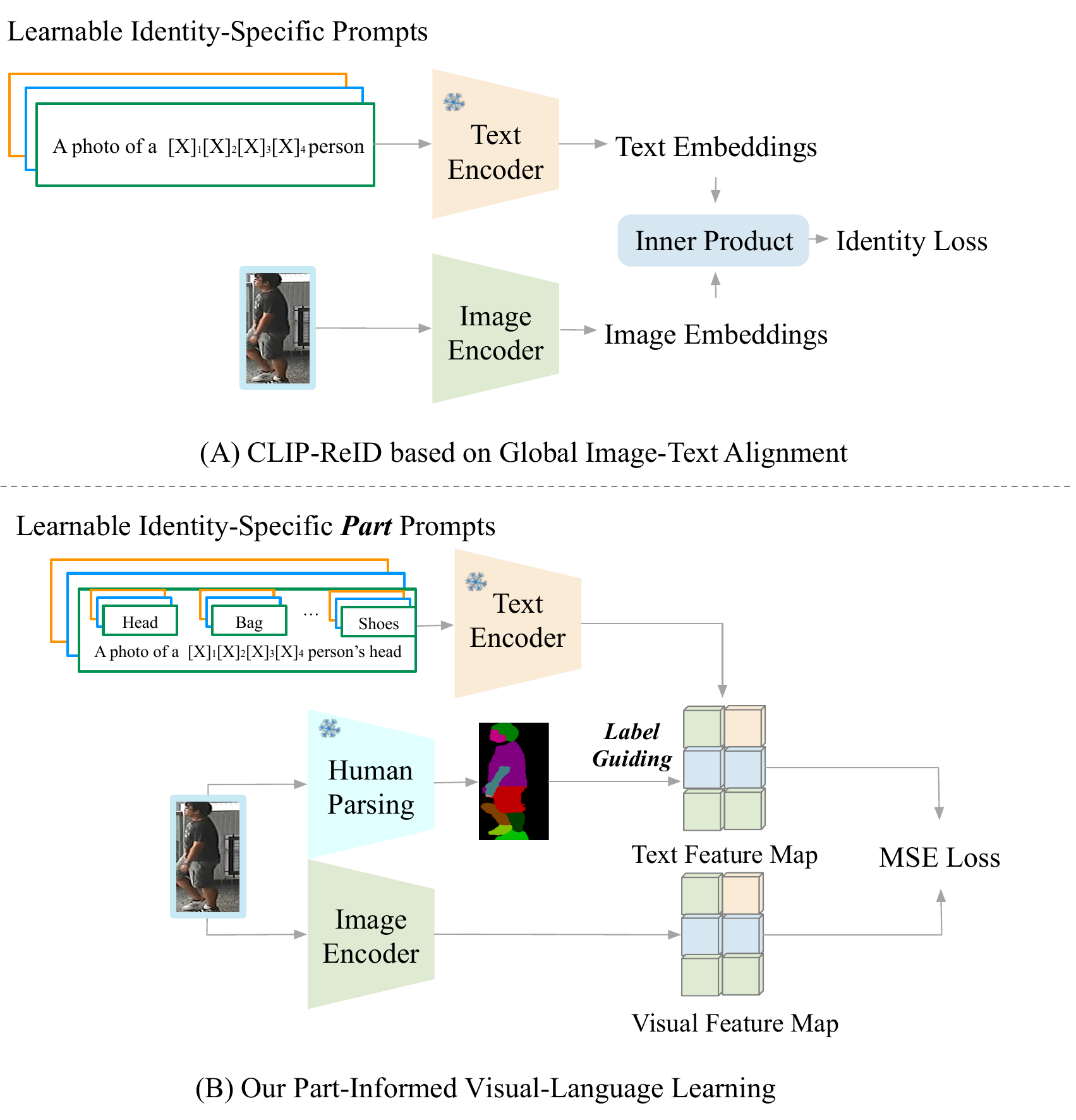}
\end{center}
    \vspace{-0.5cm} 
   \caption{Comparison of CLIP-ReID \cite{li2023clip} and our part-informed visual-language learning ($\pi$-VL) framework. (a) CLIP-ReID based on global image-text alignment. (b) Our $\pi$-VL based on pixel-level image-text alignment.}
   \vspace{-0.5cm} 
\label{fig:problem}
\end{figure}

Inspired by the recent success of visual-language models \cite{radford2021learning, li2022blip}, CLIP-ReID \cite{li2023clip} is one of the pioneer attempts that leverages natural texts to specify visual concepts beyond appearance. By tuning identity-specific text prompts \cite{zhou2022learning}, it uses text representations generated by a powerful text encoder \cite{radford2021learning} to deliver the image encoder a broader source of supervisions, leading to more discriminative global features. However, naively porting ideas from global image-text alignment may not suffice for ReID. Several studies \cite{Sun_2018_ECCV, wei2018person} have demonstrated that some non-salient details can be easily overwhelmed, raising the \textit{within-part semantic inconsistency issue} (see Fig.\ref{fig:inconsistency}). And they also revealed that introducing part-informed identity supervisions is a promising solution to this issue \cite{Sun_2018_ECCV}. 
This motivates us to ask: \textit{Is learning fine-grained body semantics as easy as global image-text alignment in ReID task?} An obstacle to addressing this issue lies in the ambiguous boundaries between different parts of the human body. While the human parsing task \cite{li2020self, ye2022invpt} has effectively tackled this problem, it introduces a new issue: \textit{supervision conflict}. Human parsing distinguishes \textit{identity-agnostic} body part semantics, whereas ReID requires \textit{identity-specific} discriminative cues. This conflict can lead to reduced feature diversity and a confused decision boundary for identity recognition.

In this paper, we address the above problems by introducing a \textbf{P}art-\textbf{I}nformed \textbf{V}isual-\textbf{L}anguage learning framework, termed $\pi$-VL, for person ReID tasks. Unlike existing works that apply parsing maps for background elimination or body alignment, we propose to construct pixel-level text prompts via human parsing, and perform per-pixel image-text alignment to enhance visual features. To alleviate the \textit{supervision conflict} problem, we combine both global-level identity labels and pixel-level parsing semantics for contrastive prompt tuning, leading to more discriminative part text embeddings. Furthermore, considering the hierarchical nature of visual backbones \cite{he2016deep}, we propose a light-weight auxiliary head to fuse multi-stage visual features and design a parsing confidence weighted alignment loss for robust semantic enhancement. It is worth noting that our $\pi$-VL is a plug-and-play and loss function-based solution, it is highly compatible with existing ReID models. Experimental results on both CNN and ViT-based backbones suggest that $\pi$-VL has the potential to be used as a universal front-end capable of handling various model architectures. 
It achieves highly competitive results, \textit{i.e.}, \textbf{91.0\%} Rank-1 and \textbf{76.9\%} mAP, on the MSMT17 benchmark, and shows consistent improvements over mainstream person ReID databases. Our contributions are summarized as follows:
\begin{itemize}
    \item We propose a part-informed visual-language learning framework, named $\pi$-VL, for person ReID. To our best knowledge, this is one of the first attempts to introduce fine-grained visual-language learning for ReID tasks.
    
    \item We present an identity-aware part-informed prompt tuning strategy based on human parsing. With this strategy, we can generate pixel-level text prompts based on both identity labels and parsing maps, strengthening the visual encoder to spot more discriminative features.

    \item We design a novel fine-grained alignment mechanism for ReID tasks. It integrates confidence scores from human parsing to weight the alignment loss, leading to a more semantically rich feature space for person image retrieval.
    
    
    \item Extensive experiments on mainstream ReID benchmarks not only demonstrate the superior performance of the proposed method, but also validate its generalization ability to various visual encoders. 
\end{itemize}

\section{Related Work}
\label{sec:related work}
\subsection{Appearance-based Person ReID}
Appearance-based ReID aims to match a target pedestrian across disjoint visible camera views at varying places and times. It is challenging to learn suitable feature representations robust enough to withstand large intra-class variations of illumination, poses, and background clutter \cite{miao2019pose, Luo_2019_CVPR_Workshops}. 


Nowadays, deep learning methods show powerful capacity of automatically extracting features from large-scale image datasets and have achieved state-of-the-art results on RGB-based person ReID tasks \cite{miao2019pose}. Building on various sophisticated CNN architectures, deep ReID models are doing exceptionally well on visual matching by learning robust cross-camera feature representations and optimal distance metrics in an end-to-end manner \cite{ye2021deep, Luo_2019_CVPR_Workshops}. 
To learn more discriminative features, part information and contextual information are also exploited in recent works \cite{wang2020high, miao2019pose}. For example, methods like PCB \cite{Sun_2018_ECCV} and MGN \cite{wei2018person} utilize hand-crafted partitioning to split feature maps into grid cells or horizontal stripes for local feature learning. Another line of researches adopt off-the-shelf pose estimation or attention module \cite{wang2020high} to extract the human part aligned features. Although these approaches have reported encouraging performance, the learned part features still lack high-level semantics under the single supervision of discrete identity labels \cite{li2023clip}.

\subsection{Visual-language Pre-training}

Over the past years, the emergence of visual-language pre-training models has led to substantial improvements to many downstream tasks \cite{radford2021learning, li2022blip}. Based on the idea of contrastive image-text alignment, CLIP  exploits a two-directional InfoNCE loss \cite{radford2021learning} to pre-train a pair of image and text encoders, leading to semantic meaningful visual representations in harmony with manually-designed text prompts. To our best knowledge, CLIP-ReID \cite{li2023clip} is the first milestone that deals with ReID tasks based on CLIP. By tuning identity-specific text prompts \cite{zhou2022learning}, it uses text representations generated by a powerful text encoder \cite{radford2021learning} to distill the image encoder, leading to more discriminative global features. However, the local part features still lack meaningful semantics under the only supervision of global identity text embeddings.




\section{Methodology}
\label{sec:method}

\subsection{Preliminaries: Overview of CLIP-ReID}
\label{sec:pre}
CLIP-ReID \cite{li2023clip} is the first milestone that applies visual-language pre-training models to appearance-based ReID. As one-hot identity labels used by ReID are meaningless to construct high-quality text prompts, it proposes a two-stage training procedure for visual-language learning.

The first training stage aims to optimize identity-specific text tokens with CLIP-style supervisions. By passing a pre-designed text description $T_{D}$= `\textit{A photo of a $[X]_{1}[X]_{2}[X]_{3}...[X]_{M}$ person.}' and the corresponding person image $I_{i}$ through a \textit{frozen} text encoder $\mathcal{T}(\cdot)$ and a \textit{frozen} image encoder $\mathcal{I}(\cdot)$ respectively, a text embedding $T_{y_{i}}$ and an image embedding $V_{y_{i}}$ could be obtained, where $[X]$ denotes a learnable text token with the word embedding dimension, $M$  is the number of learnable text tokens, and $y_{i}$ indicates the person identity label. Then, CLIP-style contrastive learning losses $\mathcal{L}_{i2t}$ and $\mathcal{L}_{t2i}$ \cite{radford2021learning} are computed to optimize $[X]_{1}[X]_{2}[X]_{3}...[X]_{M}$:
\begin{equation}
\label{eq:clip}
    \mathcal{L}_{stage1} = \mathcal{L}_{i2t} + \mathcal{L}_{t2i}.
\end{equation}

In the second training stage, the learned identity-specific text embeddings are treated as a \textit{classifier}, and the image encoder $\mathcal{I}(\cdot)$ is fully optimized under the supervision of identity loss $\mathcal{L}_{id}$ with label smoothing and triplet loss  $\mathcal{L}_{tri}$ \cite{Luo_2019_CVPR_Workshops}:
\begin{equation}
\label{eq:loss_clipreid_imtxt}
    \mathcal{L}_{i 2 t c e}=\sum_{k=1}^N-q_k \log \frac{\exp \left(V_i\cdot T_{y_k}\right)}{\sum_{y_a=1}^N \exp \left(V_i\cdot T_{y_a}\right)}
\end{equation}

\begin{equation}
\label{eq:loss_clipreid_2}
     \mathcal{L}_{stage2} = \mathcal{L}_{id} + \mathcal{L}_{tri} +  \mathcal{L}_{i 2 t c e},
\end{equation}
where $q_k$ represents the soft label in the target distribution, $N$ is the number of identities, and $i$ denotes the image index.

\subsection{The Within-part Semantic Inconsistency Issue}
\label{sec:problem}
\begin{figure}[t]
\begin{center}
   \includegraphics[width=0.7\linewidth]{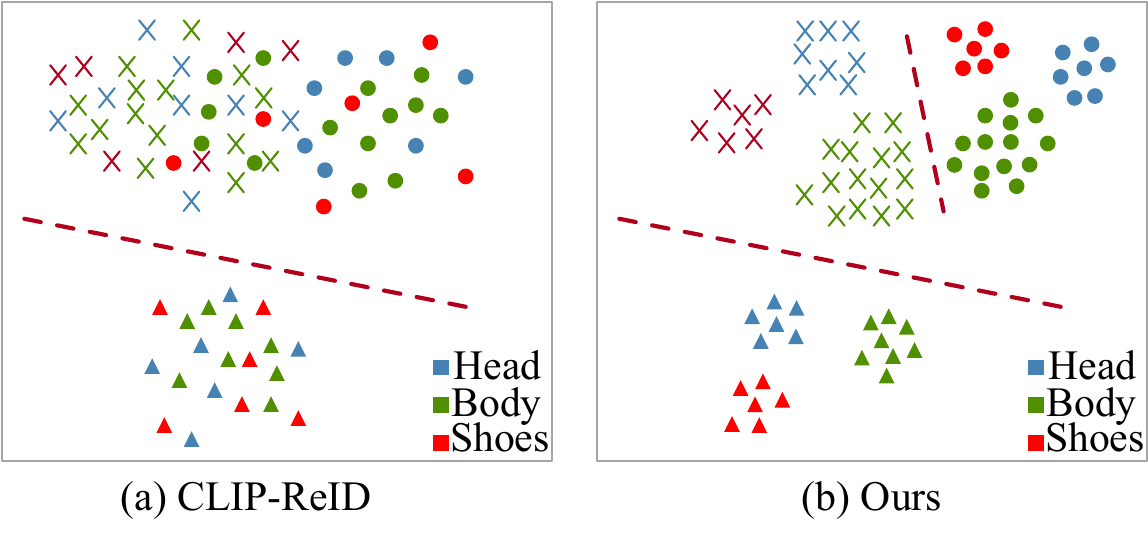}
\end{center}
    \vspace{-0.5cm} 
   \caption{Illustration of within-part semantic inconsistency. Colors indicate different body parts, symbols denote human identities, and the red dashed line represents the decision boundary for identity recognition.}
   \vspace{-0.5cm} 
\label{fig:inconsistency}
\end{figure}

CLIP-ReID is simple and effective, yet to be improved. Since the natural language supervision is limited to the whole-body scale (Eq.(\ref{eq:loss_clipreid_imtxt})), some non-salient or infrequent part features can be easily overwhelmed, and still lack high-level semantics \cite{wang2018learning}. As shown in Fig.\ref{fig:inconsistency}, we use t-SNE \cite{van2008visualizing} to visualize the pixel-level feature distributions produced by CLIP-ReID, and adopt human parsing \cite{li2020self} to assign semantic labels for each pixel. Here, colors indicate different body parts, while symbols denote human identities.

We observe that although the decision boundary of identities (the red dashed line) is generally clear, features of different body parts are hard to distinguish. Furthermore, for several confusing identities (\textit{e.g.}, identities denoted as crosses and circles), the classification boundary of their part features is even more difficult to be recognized. We term this issue as \textit{within-part semantic inconsistency}, which directly hinders the performance of person retrieval.

\begin{figure*}[t]
\begin{center}
   \includegraphics[width=0.7\linewidth]{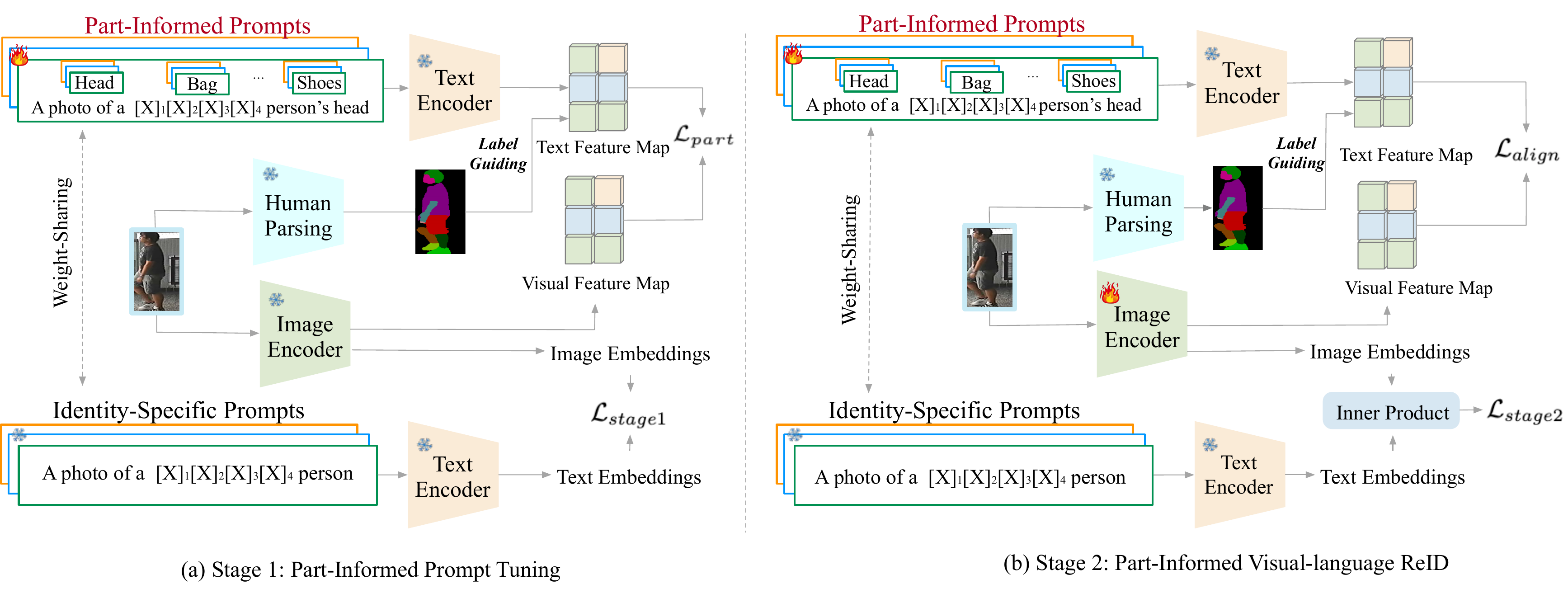}
\end{center}
\vspace{-0.6cm}
   \caption{The proposed $\pi$-VL framework. To solve the within-part semantic inconsistency issue (Section \ref{sec:problem}), it first learns identity-specific and part-informed text prompts in a coarse-to-fine manner (Section \ref{sec:prompt}). Then it leverages a hierarchical fusion-based alignment strategy (Section \ref{sec:align}) to perform fine-grained image-text alignment between part-informed text embeddings and multi-scale visual features.}
\label{fig:framework}
\end{figure*}

\subsection{Part-Informed Prompt Tuning}
\label{sec:prompt}
To address the issue of within-part semantic inconsistency, an intuitive approach is to aggregate part features that share the same semantics while separating those that are irrelevant. 
This intuition, however, further raises two questions: (1) How to identify the semantics of fine-grained part features? (2) How to design the supervision signal for part distinction?

The first question has already been answered by state-of-the-art human parsing models, which are robust to the ambiguous boundaries between different body parts. Thus, we employ a human parsing model $\mathcal{H}$ \cite{li2020self} to generate a pixel-level parsing map $P$ for person image $I_{i}$. Specifically, we follow the setup from \cite{li2020self} and classify each pixel into N (N=20) semantic categories, including `\textit{Background}', `\textit{Hat}', `\textit{Hair}', etc (see the appendix for details). This allows us to generate per-pixel text prompts based on the semantic labels and CLIP text encoder.


However, human parsing inherently introduces a new obstacle in addressing the second question. That is, human parsing only distinguishes \textit{identity-agnostic} body part semantics, whereas ReID requires learning \textit{identity-specific} discriminative cues. This conflict can suppress the diversity and discriminability of ReID features to some extent, leading to inferior performance. Inspired by \cite{zhou2022learning}, we propose a part-informed prompt tuning strategy to solve the supervision conflict issue. As illustrated in Fig.\ref{fig:framework}(a), similar to \cite{li2023clip}, we first learn identity-specific tokens with the text prompt $T_{i}$, \textit{i.e.}, `\textit{A photo of a $[X]_{1}[X]_{2}[X]_{3}...[X]_{M}$ person}', through optimizing Eq.(\ref{eq:clip}). 

But unlike \cite{li2023clip}, we reformulate the identity text prompt at the pixel level using parsing maps, thereby generating fine-grained text prompts $T_{i}^{part}$, such as, `\textit{A photo of a $[X]_{1}[X]_{2}[X]_{3}...[X]_{M}$ person's head.}', `\textit{A photo of a $[X]_{1}[X]_{2}[X]_{3}...[X]_{M}$ person's shoes.}'. 
Then we can obtain a fine-grained text embedding $t_{i}^{part}$ by passing fine-grained text prompts $T_{i}^{part}$ through the tokenizer $\mathcal{T}$:
\begin{equation}
t_{i}^{part} = \mathcal{T}(T_{i}^{part}).
\end{equation}
Then, we align the spatial resolution of visual feature maps and parsing map via downsampling, and rearrange the fine-grained text embedding based on the spital arrangement of parsing maps, leading to a `text embedding map' (see appendix for details), \textit{i.e.},
\begin{equation}
t_{i}^{full} = \text{rearrange}(t_{i}^{part}) \quad t_{i}^{full} \in \mathbb{R}^{H\times W},
\end{equation}
Next, we propose to learn our part-informed text prompts with pixel-level dense contrastive learning. Specifically, let $v_{i}^{full}$ denote the visual feature map extracted by the visual encoder, we treat pixel-wise text embedding $t_{i}^{j}\in t_{i}^{full}$ and visual embeddings $v_{i}^{j} \in v_{i}^{full}$ of the $j$-th semantic label of the $i$-th person $(t_{i}^{j}, v_{i+}^{j})$, $(v_{i}^{j}, t_{i+}^{j})$ as positive pairs, while counting the others as negative pairs $(t_{i}^{j}, v_{i-}^{j})$, \textbf{$(v_{i}^{j}, t_{i-}^{j})$}:
\begin{equation}
\small
\begin{aligned}
\mathcal{L}_{t2i}^{part}=\frac{1}{N} \sum_{j=1}^{N}&\log \frac{\exp \left(t_{i}^{j} \cdot v_{i+}^{j} / \tau\right)}{\exp \left(t_{i}^{j} \cdot v_{i+}^{j}\right)+\sum_{v_{i-}^{j}} \exp \left(t_{i}^{j} \cdot v_{i-}^{j} / \tau\right)}, \\
\mathcal{L}_{i2t}^{part}=\sum_{j=1}^{N}&\log \frac{\exp \left(v_{i}^{j} \cdot t_{i+}^{j} / \tau\right)}{\exp \left(v_{i}^{j} \cdot t_{i+}^{j}\right)+\sum_{t_{i-}^{j}} \exp \left(v_{i}^{j} \cdot t_{i-}^{j} / \tau\right)}, \\
&\mathcal{L}_{part} = \mathcal{L}_{t2i}^{part} + \mathcal{L}_{i2t}^{part},
\end{aligned}
\end{equation}

where $\tau$ is the temperature coefficient of the InfoNCE loss. During the prompt tuning process, only the learnable text tokens $[X]_{1}[X]_{2}[X]_{3}...[X]_{M}$ are optimized, while the image and text encoders are frozen (see the appendix for details).

Therefore, the loss function of the first training stage can be defined as:
\begin{equation}
\label{Eq:loss_stage1}
\mathcal{L}_{stage1’} = \mathcal{L}_{stage1} + \mathcal{L}_{part},
\end{equation}

\subsection{Part-Informed Visual-Language ReID}
With part-informed prompt tuning, we are able to generate identity-specific text embeddings with discriminative body semantics. However, for image-text alignment,  two issues remain: 1) \textit{Which resolution of visual features is suitable for supervision by language signals} and 2) \textit{How to make the alignment process robust against inevitable noises generated by off-the-shelf parsing models}?

For the first issue, downsampling operations can easily lead to information loss, causing supervision signals to become inaccurate for small-scale features. Therefore, directly imposing language supervision on all intermediate visual feature maps is not advisable. To address this, we introduce an auxiliary head that fuses multi-scale visual features to a relatively higher resolution. This allows both low- and high-level visual features to directly receive gradients from language guidance, thereby creating a more semantically meaningful feature space for person re-identification.


\label{sec:align}
\begin{figure}[h]
\begin{center}
   \includegraphics[width=0.8\linewidth]{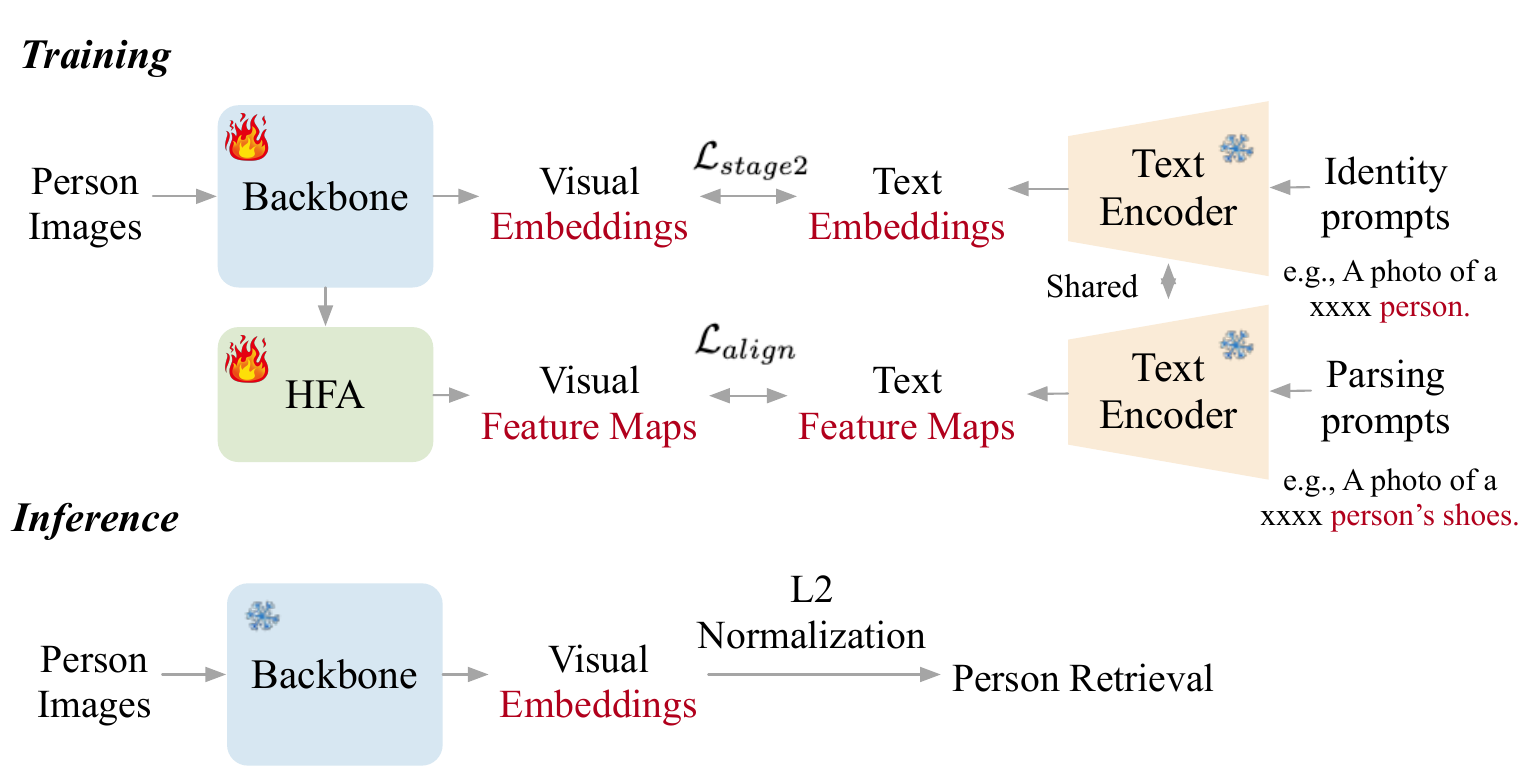}
\end{center}
\vspace{-0.2cm} 
   \caption{Illustration of the hierarchical image-text alignment strategy. We propose to fuse multi-scale features for image-text alignment.}
\label{fig:fpn}
\vspace{-0.2cm} 
\end{figure}

Figure \ref{fig:fpn} illustrates the implementation of our hierarchical visual-text alignment (HFA) strategy. For CNN-based backbones, we simply employ a lightweight feature pyramid network \cite{li2022exploring} to align the spatial and channel dimensions of output features. For ViT-based backbones \cite{dosovitskiy2020image}, we follow the design principle proposed by \cite{li2022exploring} to fuse multi-scale visual features for a plain backbone. It is worth noting that our HFA is an inference-free solution and will therefore be discarded during the inference process.

For the second issue, we empirically find that higher pixel-wise confidence scores generated by human parsing often represent accurate parsing results \cite{li2020self}. Motivated by this finding, we propose a parsing confidence-weighted alignment loss for fine-grained image-text alignment. For each pixel-level image and text feature pairs $(v_{i}^{full}, t_{i}^{full})$, we downsample the parsing confidence map $P$ to the spatial scale of visual feature maps (\textit{i.e.,} $h\times w$), and perform softmax operation to normalized it to a weight map $\hat{P}$, then we utilize $\hat{P}$ as the spatial weight of the pixel-wise mean squared error loss, \textit{i.e.},
\begin{equation}
\label{eq:loss_align}
    \mathcal{L}_{align} = \hat{P} * ||v_{i}^{full} - t_{i}^{full}||_{2}^{2}.
\end{equation}


The overall learning objective of the second training stage is a weighted summation of stage2 loss (Eq.(\ref{eq:loss_clipreid_2})) and $\mathcal{L}_{align}$ (Eq.(\ref{eq:loss_align})), defined as:
\begin{equation}
\label{Eq:loss_stage2}
    \mathcal{L}_{stage2'} = \mathcal{L}_{stage2} + \mathcal{L}_{align},
\end{equation}

During inference, we directly use the global features of the visual encoder to conduct person retrieval (Fig.(\ref{fig:fpn})).

\begin{table*}[t]
\scriptsize
\centering
\caption{Comparison with state-of-the-art methods. `repo' means our reproduced results.}
\vspace{-0.3cm}
\label{Table:SOTA}
\begin{tabular}{c|c|c|cc|cc|cc|cc}
\multirow{2}{*}{Backbone} & \multirow{2}{*}{Methods} & \multirow{2}{*}{Venues} & \multicolumn{2}{c|}{MSMT17} & \multicolumn{2}{c|}{Market-1501} & \multicolumn{2}{c|}{CUHK03} & \multicolumn{2}{c}{Occluded-Duke} \\
                          &                          &                         & mAP         & Rank-1        & mAP            & Rank-1          & mAP          & Rank-1         & mAP            & Rank-1           \\ \Xhline{1.2pt}
\multirow{5}{*}{CNN}     & PCB\cite{Sun_2018_ECCV}                      & ECCV 2018               & -           & -             & 81.6           & 93.8            & 57.5         & 63.7           & -              & -                \\
                          & MGN\cite{wang2018learning}                      & MM 2018                 & -           & -             & 86.9           & 95.7            & 67.4         & 68.0           & -              & -                \\
                          & OSNet\cite{zhou2019omni}                    & ICCV 2019               & 52.9        & 78.7          & 84.9           & 94.8            & 67.8         & 72.3           & -              & -                \\
                          & Auto-ReID\cite{quan2019auto}                & ICCV 2019               & 52.5        & 78.2          & 85.1           & 94.5            & 73.0            & 77.9              & -              & -                \\
                          & HOReID\cite{wang2020high}                   & CVPR 2020               & -           & -             & 84.9           & 94.2            & -         & -           & 43.8           & 55.1             \\
                          & CAL \cite{gu2022clothes}                      & ICCV 2021               & 56.2        & 79.5          & 87.0           & 94.5            & -         & -           & -              & -                \\
                          & LTReID\cite{wang2022ltreid}                   & TMM 2022                & 58.6        & 81.0          & 89.0           & 95.9            & -         & -           & -              & -     \\           
                        & CAJ\cite{chen2024jaccard}                     & CVPR 2024                & 44.3        & 75.1          & 86.1           & 94.4            & -         & -           & -              & -                \\ \cline{2-11} 
                          & CLIP-ReID\cite{li2023clip}                & AAAI 2023               & 63.0        & 84.4          & 89.8           & 95.7            & -         & -           & 53.5           & 61.0             \\
                          & CLIP-ReID (repo)                & AAAI 2023               & 62.5        & 84.0          & 89.2           & 95.3            & 77.8         & 81.2           & 53.2           & 60.7
                                \\
                          &\cellcolor[HTML]{EFEFEF} $\pi$-VL (ours)                    &\cellcolor[HTML]{EFEFEF}-                       &\cellcolor[HTML]{EFEFEF}\textbf{64.2}            &\cellcolor[HTML]{EFEFEF}\textbf{85.8}             &\cellcolor[HTML]{EFEFEF}\textbf{90.5}              &\cellcolor[HTML]{EFEFEF}\textbf{96.5}               &\cellcolor[HTML]{EFEFEF}\textbf{79.8}             &\cellcolor[HTML]{EFEFEF}\textbf{83.7}               &\cellcolor[HTML]{EFEFEF}\textbf{54.5}               &\cellcolor[HTML]{EFEFEF}\textbf{62.3}                 \\ \Xhline{1.2pt}
\multirow{5}{*}{ViT}      & AAformer\cite{zhu2021aaformer}                 & Arxiv 2021              & 63.2        & 83.6          & 87.7           & 95.4            & 77.8         & 79.9           & 58.2           & 67.0             \\
                          & TransReID\cite{he2021transreid}                & ICCV 2021               & 67.4        & 85.3          & 88.9           & 95.2            & -         & -           & 59.2           & 66.4             \\
                          & DCAL\cite{zhu2022dual}                     & CVPR 2022                & 64.0        & 83.1          & 87.5           & 94.7            & -         & -           & -              & -  \\           
                          & InstructReID\cite{he2024instruct}                     & CVPR 2024                & 72.4        & 86.9          & \textbf{93.5}           & 96.5            & -         & -           & -              & -                \\ \cline{2-11} 
                          & CLIP-ReID\cite{li2023clip}                & AAAI 2023               & 75.8        & 89.7          & 90.5           & 95.4            & -         & -           & 60.3           & 67.2             \\
                          & CLIP-ReID (repo)                & AAAI 2023               & 75.0        & 88.7          & 90.1           & 95.2            & 79.5         & 82.1           & 59.7           & 66.5             \\

                          & \cellcolor[HTML]{EFEFEF} $\pi$-VL (ours)                     &\cellcolor[HTML]{EFEFEF}-                       &\cellcolor[HTML]{EFEFEF}\textbf{76.9}            &\cellcolor[HTML]{EFEFEF}\textbf{91.0}              &\cellcolor[HTML]{EFEFEF}\textbf{91.3}               &\cellcolor[HTML]{EFEFEF}\textbf{97.0}                &\cellcolor[HTML]{EFEFEF}\textbf{83.0}             &\cellcolor[HTML]{EFEFEF}\textbf{84.5}               &\cellcolor[HTML]{EFEFEF}\textbf{61.4}               &\cellcolor[HTML]{EFEFEF}\textbf{69.4}                
\vspace{-0.3cm}
\end{tabular}
\end{table*}

\section{Experiments}

\subsection{Datasets and Evaluation Protocols}

We evaluate $\pi$-VL on four publicly available person ReID benchmarks, including MSMT17 \cite{wei2018person}, Market-1501 \cite{zheng2015scalable}, CUHK03 \cite{li2014deepreid}, and Occluded-Duke \cite{miao2019pose}. We follow the general ReID evaluation protocol \cite{zheng2015scalable}. The standard cumulated matching characteristics (CMC) curve and mean average precision (mAP) are used to evaluate the retrieval performance.

\vspace{-0.2cm}
\subsection{Comparisons with State-of-the-art Methods}

In this subsection, we demonstrate the effectiveness of our proposed method by comparing it with state-of-the-art ReID algorithms. The compared approaches include global visual representation learning methods \cite{zhou2019omni,quan2019auto,gu2022clothes, li2023clip, chen2024jaccard}, local visual representation learning models \cite{Sun_2018_ECCV,wang2018learning,wang2020high,gu2022clothes}, and ViT-based approaches \cite{zhu2021aaformer,he2021transreid,zhu2022dual,li2023clip,he2024instruct}. Table \ref{Table:SOTA} shows the Rank-1 accuracy and mAP of various methods across four datasets. Key observations include:



\textbf{Visual-language learning benefits RGB-based ReID.} As demonstrated by CLIP-ReID, the feature interaction between a pair of aligned image and text encoders brings substantial improvements to ReID performance. Specifically, on the MSMT17 dataset, it shows a 4.4\% absolute improvement over TransReID. On one hand, the broad supervisions provided by natural language effectively enhance the semantics of visual features. Moreover, CLIP provides a better starting point compared to the traditional ImageNet-supervised pre-training \cite{he2016deep} used in ReID models. However, CLIP-ReID only applies natural language supervision to global visual features, thereby overlooking the fine-grained semantics of part-informed features.

\textbf{Part-Informed visual-language learning matters.} 
In Fig.\ref{fig:inconsistency}, it can be observed that different body part features are difficult to distinguish using global image-text alignment (Fig. \ref{fig:inconsistency}(a)). In contrast, fine-grained visual-language learning emerges as a promising solution (Fig. \ref{fig:inconsistency}(b)). This demonstrates that our $\pi$-VL effectively alleviates the semantic inconsistency issue.
Beside, our $\pi$-VL performance highly comparable to state-of-the-arts methods on all four experimental benchmarks. Specially, we achieve \textbf{91.0\%} Rank-1 and \textbf{76.9\%} mAP for MSMT17, indicating that introducing part-informed text prompts leads to more semantically meaningful visual features.  
Notably, all the improvements are achieved under the single-query mode without re-ranking or other bells and whistles, and crucially, without increasing FLOPs or parameters. This underscores the effectiveness of $\pi$-VL

\vspace{-0.3cm}
\begin{table}[h]
\scriptsize
\centering
\caption{Evaluation of each module on the CUHK03 dataset. \\
$\mathcal{B}$: the baseline model, $\mathcal{H}$: human parsing-based prompts, $\mathcal{P}$: identity-aware part-informed prompts, $\mathcal{W}$: parsing confidence weighted alignment loss.}
\vspace{-0.2cm}
\begin{tabular}{cccc|c|c}
$\mathcal{B}$ & $\mathcal{H}$ & $\mathcal{P}$ & $\mathcal{W}$ & Rank-1       & mAP \\ \Xhline{1.2pt}
$\checkmark$ &   &   &   & 81.2 (repo)       & 77.8 (repo)           \\
$\checkmark$ & $\checkmark$ &   &   & 81.7 (\textcolor{green}{+0.5}) & 78.4 (\textcolor{green}{+0.6})     \\
$\checkmark$ &   & $\checkmark$   & & 82.5 (\textcolor{green}{+1.3}) & 78.7 (\textcolor{green}{+0.9})     \\
$\cellcolor[HTML]{EFEFEF}\checkmark$ &\cellcolor[HTML]{EFEFEF}   &\cellcolor[HTML]{EFEFEF}$\checkmark$ &\cellcolor[HTML]{EFEFEF}  $\checkmark$ &\cellcolor[HTML]{EFEFEF}\textbf{83.7} (\textcolor{green}{+2.5}) &\cellcolor[HTML]{EFEFEF}\textbf{79.8} (\textcolor{green}{+2.0})
\end{tabular}
\label{Table:ablation}
\end{table}
\vspace{-0.6cm}
\subsection{Ablation Studies}
In this section, we evaluate the effectiveness of different $\pi$-VL components on CUHK03 dataset, with the results summarized in Table \ref{Table:ablation}. Here we adopt CLIP-ReID with the ResNet-50 backbone as the baseline model. To assess the impact of part-informed visual-language learning, we first only employ the semantic labels generated by human parsing as the text prompt $\mathcal{H}$, \textit{e.g.,} `\textit{A photo of a person's head}'. Then we impose the proposed identity-aware part-informed prompts $\mathcal{P}$, \textit{e.g.,} `\textit{A photo of a $[X]_{1}[X]_{2}[X]_{3}...[X]_{M}$ person's head}'. Finally, we evaluate the effectiveness of the parsing confidence weighted alignment loss \textit{i.e.}, the full version of $\pi$-VL.

\textbf{Effectiveness of part-informed semantic labels.} As listed in Table \ref{Table:ablation}, when only exploiting the human parsing labels as text prompts to perform image-text alignment, it surprisingly yields 0.5\% gains of Rank-1 and 0.6\% enhancement of mAP. This suggests that fine-grained image-text alignment generally enriches the semantics of visual features. Unlike other part feature learning methods \cite{Sun_2018_ECCV, wei2018person}, our fine-grained image-text alignment brings no additional inference costs to $\mathcal{B}$.

\textbf{Effectiveness of identity-aware part prompts.} A limitation of parsing label-based prompts is that they only distinguish the semantics of different body parts, while ReID tasks require learning identity-related cues for person retrieval. To address this, we propose combining identity and parsing prompts for image-text alignment. As evidenced by the 3rd row of Table \ref{Table:ablation}, introducing identity labels results in a 0.8\% improvement in Rank-1 accuracy and a 0.3\% increase in mAP. These gains benefit from our \textit{coarse-to-fine} strategy, which initially learns identity-specific prompts and subsequently refines them with part-level semantics.


\textbf{Effectiveness of the parsing confidence weighted alignment loss.} In this paper, we propose a parsing confidence weighted alignment loss (Eq. \ref{eq:loss_align}) to enhance the model robustness to noises generated by human parsing. As shown in the last row of Table \ref{Table:ablation}, this loss function yields 0.6\% gains of Rank-1 and 0.6\% enhancement of mAP. This indicates that our weighted strategy effectively guides the model to focus on pixels with higher parsing confidence.
\vspace{-0.3cm}
\begin{table}[h]
\centering
\scriptsize
\caption{Generalization ability to different human parsing models.}
\vspace{-0.2cm}
\label{Table:ablation_parsing}
\begin{tabular}{c|c|cc}
Method & mIOU on \cite{hoiem2009pascal} & Rank-1 & mAP  \\ \Xhline{1.2pt}
CLIP-ReID \cite{li2023clip} (baseline)  & -   & 81.2   & 77.8 \\
SOLIDER \cite{chen2023beyond}        & 55.45   & 83.0   & 79.3 \\
\cellcolor[HTML]{EFEFEF}SCHP \cite{li2020self}      &\cellcolor[HTML]{EFEFEF}59.36   &\cellcolor[HTML]{EFEFEF}\textbf{83.7}   &\cellcolor[HTML]{EFEFEF}79.8 \\
InVPT \cite{ye2022invpt}         & 67.61   & 83.5   & \textbf{80.0}
\vspace{-0.2cm}
\end{tabular}
\end{table}

\textbf{Influence of different human parsing models.} A major concern for parsing-based ReID models is that the performance is sensitive to the quality of parsing maps. Here, we compare the ReID performance of $\pi$-VL with different human parsing models, including SOLIDER \cite{chen2023beyond}, InvPT \cite{ye2022invpt}, and SCHP \cite{li2020self}, on the CUHK03 dataset. As shown in Table \ref{Table:ablation_parsing}, our method consistently outperforms the baseline model across various parsing approaches. Besides, $\pi$-VL demonstrates high stability in ReID performance regardless of the parsing model used. This stability likely stems from conducting image-text alignment at the feature map level rather than the raw pixel level.
Furthermore, we also observe that higher MIOU on \cite{hoiem2009pascal} does not necessarily translate to a higher Rank-1 accuracy on CUHK03. This discrepancy arises because the average resolution of person images in ReID benchmarks is generally lower than that in \cite{hoiem2009pascal}, creating a domain gap that affects parsing quality. Moreover, since $\pi$-VL only involves human parsing during training, we can generate high-quality parsing maps offline, even using human annotations, to eliminate label noise in visual-language learning.

\vspace{-0.2cm}
\section{Conclusion}
In this paper, we introduce one of the first attempts to extend visual-language learning-based ReID from whole-body to fine-grained part-level. To address within-part semantic inconsistencies, we use human parsing for pixel-level labeling and design identity-aware, part-informed text prompts. This enables fine-grained image-text alignment, creating a more semantically meaningful embedding space for person ReID. We also develop a fusion module and a parsing confidence-weighted alignment loss to integrate features at different semantic levels.
Building upon these innovations, our $\pi$-VL is a plug-and-play, inference-free solution compatible with modern backbones. Experiments on CNN and ViT-based models show its superiority in general ReID tasks.

\noindent\textbf{Acknowledgments.} This work was supported by the National Key R\&D Program of China (2022YFB4500600)


\vspace{-0.2cm}
\bibliographystyle{IEEEbib}
\bibliography{icme2025references}

\end{document}